\documentclass{article}

\usepackage{graphicx}
\usepackage{amsmath}
\usepackage{multirow}
\usepackage{subfig}
\usepackage[square,sort,comma,numbers]{natbib}

\usepackage[final]{nips_2016}

\usepackage[utf8]{inputenc} % allow utf-8 input
\usepackage[T1]{fontenc}    % use 8-bit T1 fonts
\usepackage{hyperref}       % hyperlinks
\usepackage{url}            % simple URL typesetting
\usepackage{booktabs}       % professional-quality tables
\usepackage{amsfonts}       % blackboard math symbols
\usepackage{nicefrac}       % compact symbols for 1/2, etc.
\usepackage{microtype}      % microtypography

\title{Bio-Inspired Multi-Layer Spiking Neural Network Extracts Discriminative Features from Speech Signals}

\author{
Amirhossein Tavanaei \ \texttt{and} \ Anthony Maida\\
 \\
The Center for Advanced Computer Studies, Bio-inspired AI Lab\\
The School of Computing and Informatics\\
University of Louisiana at Lafayette\\
Lafayette, LA 70504, USA\\
\texttt{\{tavanaei,maida\}@louisiana.edu}
} 

\usepackage{fancyhdr} 

\makeatletter
\def\ps@pprintTitle{\def@oddfoot{}}
\makeatother

\begin{document}
\pagestyle{fancy}
\maketitle

\begin{abstract}
%Hierarchical feature discovery using multi-layer neural networks (NNs) has performed remarkably in pattern recognition. In contrast to conventional neural networks (NNs), 
Spiking neural networks (SNNs) enable power-efficient implementations due to their sparse, spike-based coding scheme. 
This paper develops a bio-inspired SNN that uses unsupervised learning to 
extract discriminative features from speech signals, 
which can subsequently be used in a classifier. 
The architecture consists of a spiking convolutional/pooling layer followed by 
a fully connected spiking layer for feature discovery. 
The convolutional layer of leaky, integrate-and-fire (LIF) neurons
%maps the signal frequencies to a higher dimensional feature space representing 
represents primary acoustic features.
%using the shifted difference-of-Gaussian (DoG) filters.
The fully connected layer is equipped with a probabilistic spike-timing-dependent plasticity learning rule. 
This layer represents the discriminative features through probabilistic, LIF neurons.
%implementing a winners-take-all competition to extract independent feature values.
%Our %intermediate 
%results show that the convolutional layer supports a multi-layer architecture. 
To assess the discriminative power of the learned features,
%The features obtained from the SNN are classified by 
they are used in a hidden Markov model (HMM) for spoken digit recognition. 
The experimental results show performance above 96\% that compares favorably with popular statistical feature extraction methods.
%such as Mel-scaled Frequency Cepstral Coefficient (MFCC).
Our results provide a novel demonstration of unsupervised feature acquisition in an SNN\@.
\\

\noindent
\textbf{Keywords:} Bio-inspired multi-layer framework, spiking network, speech recognition, unsupervised feature extraction.
\end{abstract}

\section{Introduction}
\label{sec:intro}

Multi-layer neural network (NN) learning extracts signal features by using a hierarchy of non-linear elements~\cite{lecun2012}. 
Such networks extract increasingly complex, discriminative, and independent features 
and serve as the basis for deep learning architectures~\cite{bengio2009,lecun2015}. 
The power of a multi-layer neural architecture can be used for a challenging problem 
like automatic speech recognition (ASR). Speech signal characteristics are highly variable in time and frequency. The feature extraction method must convert the speech signal to discriminative features 
to support the recognition task. 
Recently, multi-layer neural architectures have outperformed previous ASR models.
Examples are convolutional neural networks (CNNs)~\cite{abdel2014}, deep CNNs~\cite{sainath2013}, deep NNs~\cite{hinton2012}, and deep recurrent NNs~\cite{graves2013}. 
Despite the high performance of these architectures, 
training and then using them is expensive from a power consumption viewpoint. 
The quest to meet a power-efficient framework can be accomplished by a spike-based neuromorphic platform. 
Spiking neural networks (SNNs)~\cite{maass1997,ghosh2009,kasabov2013dynamic} provide an appropriate starting point for this. After introducing
the power of SNNs as a third generation neural network~\cite{maass1997}, a number of studies have concentrated
on biologically motivated approaches for pattern recognition~\cite{diehl2015unsupervised,kheradpisheh2016bio,bengio2017stdp}. An SNN architecture consists of spiking
neurons and interconnecting synapses undergoing spatio-temporally local learning in response to the stimulus presentation.

Word recognition functionality in the auditory ventral stream of the human brain is enabled by a multi-layer, 
biological SNN. 
Multi-layer SNNs have performed comparably with conventional NNs in visual pattern recognition tasks~\cite{masquelier2007,wysoski2008,beyeler2013}. 
Also, SNNs have been employed for the spoken word recognition tasks such as using synaptic weight association training~\cite{wade2010}, extracting spike signatures from speech signals~\cite{tavanaei2017}, and Gaussian mixture model (GMM) implementation using SNNs~\cite{tavanaei2016a}. 
Representing speech signal features through temporal spike trains has attracted much interest. 
Verstraeten et al.\@ developed a high performance reservoir-based SNN for spoken digit recognition problem~\cite{verstraeten2006}. The reservoir is not trained and it maps the inputs to a higher dimensional space to be linearly separable. Dibazar et al. proposed a feature extraction method using a dynamic synaptic neural network for isolated word recognition~\cite{dibazar2003}. In another study, Loiselle et al. utilized the cochlear Gammatone filter bank to generate spike trains in French spoken word recognition based on rank-order coding with an SNN~\cite{loiselle2005}. However, these studies did not develop a multi-layer SNN to extract discriminative speech characteristics in a hierarchy of feature discovery neural layers. 

This paper develops a multi-layer spiking neural network consisting of a convolutional layer, a pooling layer, 
and an unsupervised feature-discovery layer. 
The convolutional layer extracts primary acoustic features through the spike trains emitted from the feature map neurons. The pooling layer reduces size of the feature maps by a novel max-pooling operation based on Mel-scaled patch sizes. 
The feature discovery layer receives presynaptic spikes from the pooling layer and undergoes learning via a probabilistic spike-timing-dependent plasticity (STDP) rule. 
Neurons in this layer represent the final features extracted from a speech signal. During the training phase, 
weights connecting the pooling layer and the feature discovery layer are adjusted based on the speech signal characteristics. 
The features extracted by the SNN are used for training and evaluating a hidden Markov model (HMM)~\cite{rabiner1989} to handle temporal behavior of the speech signal. 
The proposed network offers a novel way to discover discriminative features for spoken digits while implementing a 
bio-inspired model, potentially enabling low-power consumption on appropriate hardware~\cite{cao2015}.       

\section{Network Architecture}
\begin{figure}
\centering
\subfloat[]{
\includegraphics[scale=0.6]{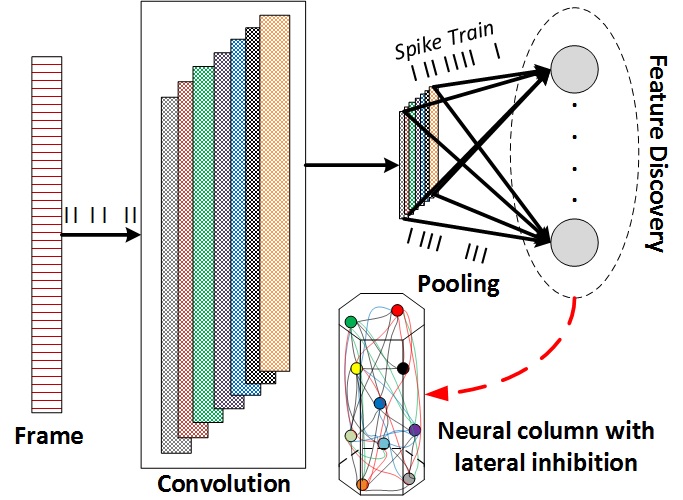}
\label{fig:model}
}
\quad
\subfloat[]{
\includegraphics[scale=.65]{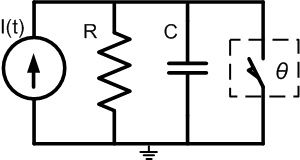}
\label{fig:RC}
}
\caption{(a): Multi-layer SNN architecture. Each rectangle shows a frame's feature vector in each layer. The convolutional layer produces seven 1-D feature maps of spiking neurons. (b): RC circuit implementing the LIF neuron with $R=1\ \Omega$, $C=1$ mF, and threshold, $\theta=\theta^{\mathrm{conv}}$.}
\end{figure}

Our multi-layer SNN, shown in Fig.~\ref{fig:model},  consists of two spiking neural layers which follow an input layer. 
%Fig.~\ref{fig:model} shows the model architecture. 
The input layer's stimulus is the magnitude Fourier transform of a speech frame ($M$ values). This layer converts the stimuli to spike trains using a Poisson process over $T=40$ ms. 
The convolutional layer maps the presynaptic spike trains to a higher dimensional feature space using convolutional filters. 
The pooling layer reduces the feature map dimension, according to Mel-scaled frequencies, 
and then transfers the new spike trains to the fully connected feature discovery layer for unsupervised
training. 
After training,
the neuronal activities in the feature discovery layer represent the frame's feature vector
that may be input to a classifier (e.g., an HMM classifier).

\section{Spiking Convolutional Layer}
The convolutional layer receives presynaptic spike trains from the input layer and generates $D$ independent
feature maps. 
Each feature map contributes to a convolutional filter acquired by shifted $p$-point difference-of-Gaussian (DoG) ($p=7$ in this network) filters. Therefore, seven shifted DoG filters are convolved with the presynaptic spike trains to generate $D=7$ feature maps. An LIF neuron in a feature map integrates the convolution results for a particular signal patch ($p=7$ points) over $T=40$ time steps and fires a spike when reaching threshold. The neuron's membrane potential, $U(t)$, is reset to the resting potential (zero) upon firing. Eq.~\ref{eq:lif} calculates the membrane potential of neuron $m$ in feature map $k$ at time $t$.

\begin{subequations}
\label{eq:lif}
\begin{align}
&\frac{dU_m^k(t)}{dt}+U_m^k(t)=I_m^k(t)\label{eq:conv1}\\
&\mathrm{if} \ \ U_m^k(t) \geq \theta^{\mathrm{conv}}, \mathrm{then} \ \ U_m^k(t)=0 \label{eq:conv2}
%I_\mathrm{syn}(t) &= \sum_{i=1}^{N_\mathrm{rec}} G_\mathrm{syn}(t-t_{i}^\mathrm{f}) (E_\mathrm{syn} - V(t))\label{eq:conv1}\\
%         &= G_\mathrm{syn}^\mathrm{tot}(t) (E_\mathrm{syn} - V(t)) \label{eq:conv2}
\end{align}
\end{subequations}
$I_m^k$ is computed by
\begin{equation}
I_m^k(t)=\sum_{i=1}^p W^k_i\cdot s_{m,i}(t)
\end{equation}
The threshold, $\theta^{\mathrm{conv}}$, is set to 0.4 in the experiments. The $I_m^k(t)$ value determines the convolution of the filter, $W^k$, and the presynaptic spike train, $s_m$, in $T=40$ time steps. This value is considered as an injected current in the RC circuit (Fig.~\ref{fig:RC}), which simulates the LIF neuron. The length of a feature map is same as the input vector.

\section{Mel-Scaled Pooling Layer}
The human hearing mechanism resolves frequencies non-linearly with higher resolution at lower frequencies. The Mel scale makes the features close to the hearing model. In the pooling layer, the stride value (pooling window size) is computed based on the Mel-scaled bandwidths. Eq.~\ref{eq:mel} converts the frequency scale to the Mel scale. Same stride value $l=2$ is used for the frequencies in the range 0 to 1000 Hz. Then, the stride values increase corresponding to the feature frequency. The number of pooling windows ($N$) is analogous to the number of Mel-spaced filter banks (20-40) selected for computing the Mel-scaled frequency cepstral coefficients (MFCCs).
The pooling layer selects a neuron with the highest activity in a pooling window (max pooling). Activities of the neurons in a feature map can be characterized by their spike rate. This layer summarizes and extracts effective signal characteristics.

\begin{equation}
\label{eq:mel}
f_{mel} = 1125 \ln \big (1+\frac{f}{700}\big )
\end{equation}

The new pooling strategy and the Mel-scaled bandwidths can also be employed by conventional CNNs for extracting acoustic features.
\section{Feature Discovery Layer}
The feature discovery layer, containing $H$ neurons and equipped with lateral inhibition, 
is fully connected to the pooling layer.
%The fully connected SNN layer is known as feature discovery layer which 
It learns and extracts the signal features based on the provided Mel-scaled pooling features. 
This layer conveys $D\times N$ spike trains emitted from the pooling layer to $H$ probabilistic LIF neurons.
The feature discovery layer is shown in the last layer of Fig.~\ref{fig:model}.
\subsection{Neuron Model}
The stochastic LIF neurons in the feature discovery layer fire based on both their membrane potential 
and their firing posterior probability (imposed by a softmax function). 
A neuron fires if its membrane potential reaches the threshold, $\theta^{\mathrm{h}}$, 
(following Eq.~\ref{eq:lif}) and its 
posterior probability is greater than 0.5. In the experiments, $\theta^{\mathrm{h}}$ is set to 3.
The posterior probability of neuron $h$ firing at time $t$ given input spike vector $\mathbf{y}_t$ 
and weights $W$ is given by Eq.~\ref{eq:posterior}. The posterior probability determines a softmax function in which the mutual exclusivity is not enforced. 
This neuron model implements a winners-take-all competition (inhibition) in the feature discovery layer. 
Although there are no direct inhibitory connections between neurons, 
the softmax function imposes an overall inhibition upon the neural column extracting independent features. 
The neural column shown in Fig.~\ref{fig:model} illustrates that because of the lateral inhibitions, only a small subset of the neurons can fire at the same time. 
The inhibition constrains the model to extract independent features in response to the input. 

\begin{equation}
\label{eq:posterior}
P(z_h=1|W, \mathbf{y}_t) = \frac{e^{W_h^T \cdot \mathbf{y}_t}}{\sum_{j=1}^H e^{W_j^T \cdot \mathbf{y}_t}}
\end{equation}
\subsection{Learning}
Spike-timing-dependent plasticity (STDP) is a learning rule used in SNNs that increases the weight of a synapse if the presynaptic spike occurs briefly before the postsynaptic spike (LTP) and decreases the synaptic weight otherwise (LTD)~\cite{dan2006}. We use a probabilistic version of the STDP rule provided by Tavanaei et al.~\cite{tavanaei2016b} (Eq.~\ref{eq:stdp}) to learn the final acoustic features.
\begin{equation}
\label{eq:stdp}
\Delta w_{hi} = 
\begin{cases}
&a^+e^{-w_{hi}}, \ \ \ \mathrm{if} \ i \ \mathrm{fires \ in } \ [t-\epsilon , t]\\
&-a^-, \ \ \ \ \ \ \  \ \mathrm{otherwise.}
\end{cases}
\end{equation}
In the above,
$a^+=10^{-3}$ and $a^-=0.75\times 10^{-3}$ are, respectively, LTP and LTD learning rates.
When a postsynaptic neuron $h$ fires, STDP is triggered for the incoming weights. 
LTP occurs if the presynaptic neuron $i$ fires briefly (e.g. within $\epsilon=5$ ms) 
before $h$ fires, while the weight change magnitude is controlled by the current weight. 
Weights remain in the range (0, 1). The probability derivations and justifications are given in~\cite{tavanaei2016b}. 
Weights are initialized by sampling uniformly in the range (0, 1). 

\subsection{Final Feature Representation}
A speech frame is built from $H$ neurons in the feature discovery layer. 
The elements of the vector are the neurons' activities in response to received spike trains via trained weights, $W$. 
%In this paper 
More specifically,
the accumulated membrane potentials of the neurons are used for the feature values. 
Samples with $H$ attributes obtained by the feature discovery layer are then used to train the HMM classifier. 

\section{Experiments and Results}
This investigation uses the proposed SNN to discover and extract discriminative features from the spoken digits 0 to 9 selected in the Aurora dataset~\cite{aurora2000} that is recorded at the sampling rate of 8000 Hz. Each speech signal (spoken digit) is divided into 25 ms frames with 50\% temporal overlap, resulting in about 60 frames per digit. 
The frames are converted to the frequency domain by using the magnitude of the Fourier transform.
Each frame is represented by 100 sample values (0 to 4000 Hz). 

\begin{figure}
\centering
\subfloat[]{
\includegraphics[scale=0.42]{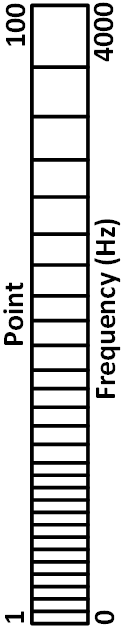}
\label{fig:pooling}
}
\quad
\subfloat[]{
\includegraphics[scale=0.37]{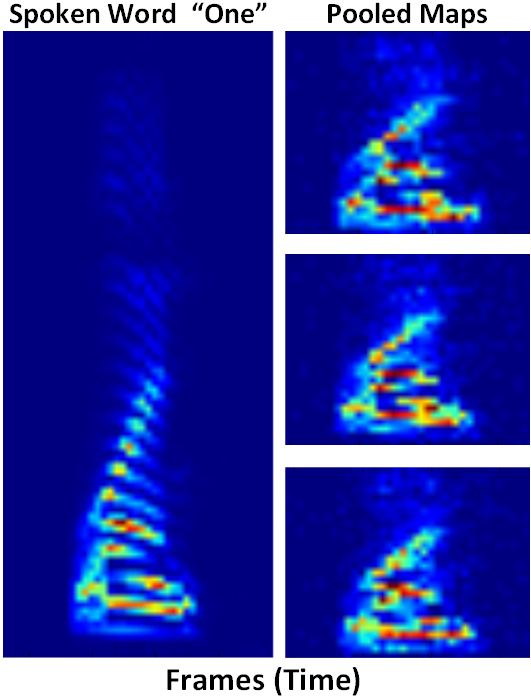}
\label{fig:fmap}
}
\quad
\subfloat[]{
\includegraphics[scale=0.38]{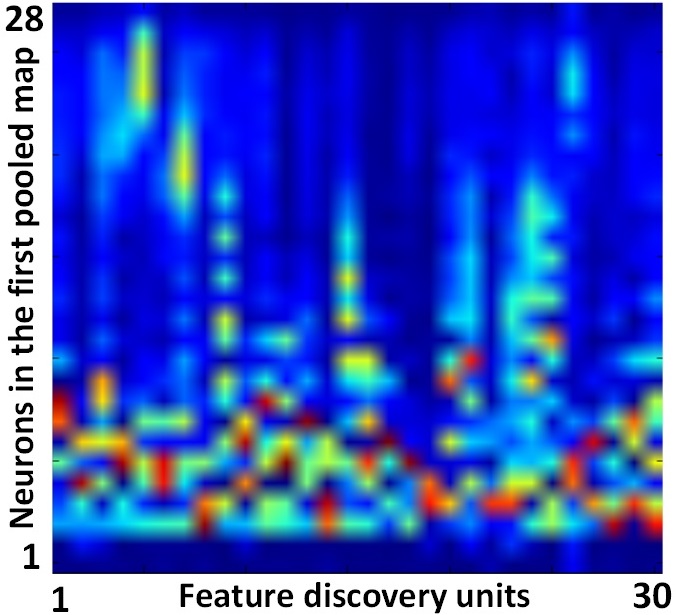}
\label{fig:weights}
}
\caption{(a): Mel-scaled stride values spanning 0-4000 Hz. 
(b): Neural spike rates of three selected pooled maps extracted from the spoken digit `one'. 
%The neuron activity corresponds to the neuron's spike rate. 
Neurons emit zero (dark blue) to 39 (dark red) spikes.
(c): Synaptic weights of the feature discovery layer connecting the neurons in a pooled map to 30 LIF neurons. The synaptic weights fall in the range 0 (dark blue) to 1 (dark red).}
\end{figure}

DoG convolutional filters are obtained by the difference of two Gaussian functions with variances $\sigma^2=1$ and $\sigma^2=6$ (Eq.~\ref{eq:dog}). 
Seven filters produce $D=7$ feature maps containing 100 spiking neurons. 
Each map is converted to a pooled map by the max-operation over the Mel-scaled pooling windows. 
The pooling windows are shown in Fig.~\ref{fig:pooling} with the stride values 2 through 10. 
In the experiments, we use 28 values to represent each speech frame. The stride values for the sample points 1 to 26 (0 to 1040 Hz) are set to 2 (13 feature values). The last stride value is 10. 
Fig.~\ref{fig:fmap} shows neural spike rates of three selected pooled maps 
extracted from the spoken digit `one'. 
The pooled layer conveys the speech frame characteristics with only $7\times 28=196$ values.    
\begin{equation}
\label{eq:dog}
\begin{aligned}
\mathrm{DoG}=\mathcal{N}(x,\mu=0,\sigma^2=1)-\mathcal{N}(x,\mu=0,\sigma^2=6), \\
x=\{-3,-2,...,3\} \ \ \ \ \ \ \ \ \ \ \ \ \ \ \ \ \ \ \ \ \ \ \ \ \ 
\end{aligned}
\end{equation} 

The feature discovery layer extracts the final feature vectors for the speech frames using 
its trained synaptic weights. Training the network involved cycling through a dataset of 1000 spoken digits for 10 iterations. 
Fig.~\ref{fig:weights} shows an example
synaptic weight set connecting a pooled map and the feature discovery 
neurons. 
Most neurons show higher resolution at lower frequencies. 
The feature vectors extracted from the temporal frames are used to train and evaluate the HMM. 
In the first experiment, $H=30$ feature discovery neurons (as feature vector dimension) were used to train 12 HMMs with $\{5,10,15\}$ states and $\{4,8,16,32\}$ GMMs embedded in each state. The testing set was sampled from the spoken digits that were not used for the training. Table~\ref{tab:acc} shows the accuracy rates reported for the spoken digit recognition task. 
Performances higher than 95\% were achieved by the HMMs with 10 and 15 states and 32 GMMs. These accuracy rates show that the SNN, which uses unsupervised learning, has extracted discriminative features from the speech signals.
\begin{table}[]
\centering
\caption{Accuracy rates of the HMMs trained and evaluated by the features extracted by the multi-layer SNN. $S$: number of states. $G$: number of GMMs.}
\label{tab:acc}
\begin{tabular}{|l|l|l|l|l|}
\hline
\textbf{HMM Variations} & \multicolumn{1}{l|}{\textit{\ G = 4\ \ }} & \multicolumn{1}{l|}{\textit{\ G = 8\ \ }} & \multicolumn{1}{l|}{\textit{\ G = 16\ }} & \multicolumn{1}{l|}{\textit{\ G = 32\ }} \\ \hline
\textit{S = 5}          & \ 62.88          & \ 75.59          & \ 85.95           & \ 90.97           \\ \hline
\textit{S = 10}         & \ 71.91          & \ 87.63          & \ 92.31           & \ 95.32           \\ \hline
\textit{S = 15}         & \ 84.95          & \ 91.30          & \ 96.32           & \ 96.32           \\ \hline
\end{tabular}
\end{table}

\begin{figure}
\centering
\includegraphics[scale=0.65]{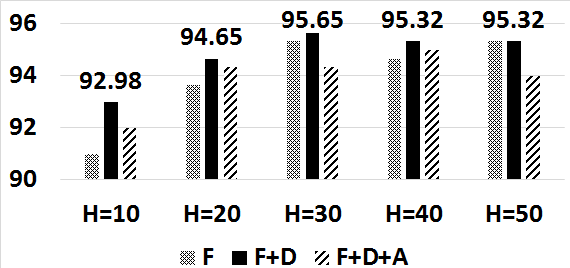}
\caption{Performance of the networks with 10 through 50 spiking neurons in the feature discovery layer for different sets of features. $\Delta$ (D) and $\Delta \Delta$ (A) features are concatenated to the extracted features (F).}
\label{fig:acc2}
\end{figure}

The next experiment evaluates: 1) the number of extracted features, $H$, (number of neurons in the feature discovery layer); and, 2) applying $\Delta$ (differential, \textbf{D}) and $\Delta \Delta$ (acceleration, \textbf{A}) features to use more dynamic information. The HMM with ten states and 32 GMMs is selected for this experiment. Fig.~\ref{fig:acc2} shows that the model with $H=30$ probabilistic neurons in the feature discovery layer outperforms the other architectures. Additionally, the $\Delta$ features improved the accuracy rate. However, the $\Delta \Delta$ features did not improve performance in the networks with more than $H=20$ output neurons. 
This could happen because the HMM might be overtrained by the high dimensional feature vectors.       

Our SNN extracted discriminative features that were classified at 96\% accuracy. 
This accuracy rate is comparable with~\cite{wade2010} (95.25\% accuracy) that uses synaptic weight association training in a neural layer. Although the reservoir approach~\cite{verstraeten2006} has performed better (near perfect) on spoken digit recognition, the question of training a reservoir to scale to larger problems remains unresolved. 
Additionally, our multi-layer SNN provides an environment to support a deep learning architecture 
in a spiking platform that compares with conventional deep networks.
Table~\ref{tab:compare} shows the performance of spoken digit recognition models using spiking neural networks and recent conventional approaches. The proposed model compares favorably with the high performance methods, such as conventional CNN (with $>99$\% accuracy), while using a power-efficient, biologically plausible framework. 
\begin{table}[]
\centering
\caption{Spoken digit recognition models using SNNs (Top) and non-SNN (Bottom) architectures evaluated on the Aurora~\cite{aurora2000}, TIDigits~\cite{tidigits}, and TI46~\cite{ti46} spoken digit datasets. Aurora is based
on a version of the TIDigits dataset, but downsampled at 8 kHz, and is roughly comparable to the TI46 dataset.}
\label{tab:compare}
\begin{tabular}{|l|l|l|l|}
\hline
\multicolumn{1}{|c|}{\textbf{Model}} & \multicolumn{1}{c|}{\textbf{Method}}       & \multicolumn{1}{c|}{\textbf{Dataset}} & \multicolumn{1}{c|}{\textbf{Acc. (\%)}} \\ \hline
Wade et al.~\cite{wade2010}                          & Synaptic weight association (SNN) & TI46                                  & 95.25                                  \\ 
Tavanaei et al.~\cite{tavanaei2017}                      & Single layer SNN and SVM classifier        & Aurora                                & 91                                     \\ 
Verstraeten et al.~\cite{verstraeten2006}                   & Reservoir computing (LSM) & TI46                                  & \textgreater97.5                       \\ 
\textbf{Our Model}                            & Spiking CNN and HMM classifier             & Aurora                                & 96                                     \\ \hline
Dao et al.~\cite{dao2014}                           & Structured sparse representation           & Aurora                                & 95                                     \\ 
Van Doremalen et al.~\cite{van2008spoken}                     & Hierarchical Temporal Memory               & TIDigits                              & 91.6                                   \\ 
Neil et al.~\cite{neil2016effective}                     & MFCC and Deep RNN              & TIDigits                              & 96.1                                   \\ 
Groenland et al.~\cite{groenland2016efficient}                     & CNN and deep shifting methods              & TI46                                  & \textgreater99                         \\ \hline
\end{tabular}
\end{table}

\section{Conclusion}
A multi-layer SNN was developed to give a novel demonstration of unsupervised
feature extraction and discovery for speech processing.
%in this investigation. 
The SNN receives input in the form of spike trains from incoming speech signals, 
extracts Mel-scaled convolutional features,
and then
extracts higher-level discriminative features. 
The convolutional layer extracts primary acoustic features to make the network layer stack-admissible for the next  layers. 
The pooling layer summarizes the primary features such that the lower frequencies had higher resolution using Mel-scaled pooling windows. As far as we know, this is a novel feature of the network. 
The last layer is the feature discovery layer equipped with the probabilistic STDP 
learning rule.
It is trained to discover more complex, independent features using stochastic LIF neurons. 
The classification results reported by the HMM (with 96\% accuracy for the spoken digit recognition problem) showed that the network architecture can discover effective, discriminative acoustic features. 
Furthermore, the multi-layer network uses sparse, spike-based coding which can be implemented by 
appropriate, power-efficient VLSI chips.
 
The initial promising results obtained by 
unsupervised learning in this model offer promise of scaling to larger problems, such as
developing a deep spiking network for the large vocabulary speech recognition (LVSR) problem. Our future work concentrates on a multi-layer SNN for feature extraction followed by a spike-based sequential feature classifier.  

\bibliographystyle{unsrt}

\bibliography{amir}

\end{document}